# Toward Mutual Trust Modeling in Human-Robot Collaboration

BASEL ALHAJI

Simulation science center Clausthal-Göttingen, TU Clausthal

ANDREAS RAUSCH

Institute for software and system engineering, TU Clausthal

MICHAEL PRILLA

Department of informatics, TU Clausthal

The recent revolution of intelligent systems made it possible for robots and autonomous systems to work alongside humans, collaborating with them and supporting them in many domains. It is undeniable that this interaction can have huge benefits for humans if it is designed properly. However, collaboration with humans requires a high level of cognition and social capabilities in order to gain humans' acceptance. In all-human teams, mutual trust is the engine for successful collaboration. This applies to human-robot collaboration as well. Trust in this interaction controls over- and under-reliance. It can also mitigate some risk. Therefore, an appropriate trust level is essential for this new form of teamwork. Most research in this area has looked at trust of humans in machines, neglecting the mutuality of trust among collaboration partners. In this paper, we propose a trust model that incorporates this mutuality captures trust levels of both the human and the robot in real-time, so that robot can base actions on this, allowing for smoother, more natural interactions. This increases the human's autonomy since the human does not need to monitor the robot's behavior all the time.

## 1 INTRODUCTION AND BACKGROUND

Robots and autonomous systems have become essential in our daily life. With their ever-increasing intelligence, robots have started to shift from tools to be used by humans to teammates working autonomously alongside them. By combining the complementary abilities of both, a team can achieve more than any single entity can achieve on its own. This requires that team members properly rely on and trust each other's capabilities.

Imagine the following scenario: a human is working with a collaborative robot on disassembling an electric car battery. The robot has a screwdriver as an end-effector, and it can autonomously recognize and loosen some kind of screws. The human task is to make sure that the battery is correctly located with correct orientation. If the human does not trust and rely on the robot loosening the screws for him/her, s/he should perform this task on his/her own, which will prevent a potential higher performance and increase the human workload. On the other hand, if the human overly trusts the robot, s/he might expect it to safely detect and loosen screws with special shapes or conditions (e.g. rusty). This might be very risky without the human intervening.

Inappropriate reliance, as described in the previous example, is a well-known problem in human-automation and human-robot collaboration (HRC) because it may lead to undesired consequences. This problem is caused by the inappropriate trust the human (as "trustor") has toward his/her robotic partner (as "trustee") [1], [2].

Trust within a team is a crucial factor for fruitful collaboration, especially when the tasks of the team members are interrelated and interdependent [3] and when they include risks [4]. Taking into consideration that "well-calibrated **mutual** trust among team members" [5] is a cornerstone of teamwork, in HRC both agents should have the ability to trust each other for successful collaboration. Using our illustrative example, if the human mistakenly puts the battery in a wrong orientation, serious consequences can happen, since the robot may hit the modules of the battery with its screwdriver end-effector. Therefore, the robot should be able to detect (un)trustworthy behaviors of the human as well and build trust on human actions.

However, most research in this area has been done on the trust of humans toward robots or other autonomous agents. Little is known about trust of robots toward the humans and, consequently, there are no models of mutual trust in HRC. The aim of our work is to develop a model of trust that is implementable on the robot, can capture the human trust in the robot in real-time, and that incorporates the mutuality of trust. That means, the human and the robot are trustors and trustees at the same time.

### 1.1 Modeling and Measuring Trust

Trust is not a simple univariate quantity [6] but rather a complex and multi-dimensional construct [7]. Many researchers from different domains put efforts in modeling and identifying the dimensions of human trust in technology [8], computers [9], automation [2] and robots [7], [10], aiming at developing methods in which trust can be quantified. Most of the existing instruments to measuring trust, however, rely on post-hoc questionnaires, which cannot be used for estimating the level of trust during the collaboration. Similar to [11], we believe that trust models must be implementable on the robot in order to recognize the current trust level the human has and use it in the decision making process of the robot to act accordingly.

### 1.2 Trust in human-robot collaboration

As mentioned earlier, human trust has a very complex construct and consists of many different dimensions. Aiming at quantifying the effects of different trust antecedents, a meta-analysis over the existing literature on human-robot trust has been conducted by Hancock et.al. [10]. The result of their analysis indicates that robot-related, performance-based factors have the strongest influences on trust in human-robot interaction.

Based on Hancock's results, some researchers simplified the trust construct and implemented computational models of trust based on performance (originally identified by [12]) with different measures of performance depending on the task [13]–[15]. The main drawback of these models is that they do not consider all trust dimensions relevant to their use cases which oversimplifies the concept of trust and makes the accuracy and validity of these models questionable.

Further, although robot trust in human partners can be critical in collaboration settings, where partners actions are interdependent, robot trust in its human partner is rarely addressed in the existing literature.

Vinanzi et al. propose an artificial cognitive architecture that can estimate the trustworthiness of the human partner [16] as a source of information. However, they did not consider physical interactions. In a physical collaboration setting, as stated in our illustrative example, trust models should also include many additional inputs, such as the human movements and actions. These are essential for the robot to form its reaction flawlessly. Other researchers consider trust of robots in humans by implementing the same model for both partners [13]–[15]. Consequently, these models handle human trust and robot trust equally and do not take the differences between the human and the robot into account.

Here, we argue that a robot should develop its own sense of trust toward its human partner. It helps the robot to adapt its behavior according to the current trustworthiness signals of the human, which increases the human safety and enhances the quality of interactions. It may also increase the autonomy of the human and reduce the workload. We should emphasize here that our motivation for this is **not** judging the human behavior and performance but rather to calculate for some unwanted human errors that might occur during the operation.

## 2  OPEN ISSUES

The mutual nature and formation of trust in teamwork is absent in the existing literature of HRC. Mostly, the human is the trustor, and the robot is the trustee. This may imply that either the robot has unlimited and unconditional trust toward the human partner or that it does not trust the human at all. Unconditional trust, however, may have negative effects on performance [17] and safety as well. Having no trust means verifying every human action and potentially slowing down cooperation. For this reason, we argue that trust models should be bidirectional and include both partners' trust in each other. This results in at least three open issues that we tackle in our work: firstly, there is a lack of knowledge on the relevant dimensions of trust in human-robot physical collaboration where partners work in close vicinity on joint tasks. Secondly, there is hardly any detailed analysis of robot trust into a human partner, although this might strongly affect the success and failure of the team. Thirdly, the influences of the robot's trust on the human behavior and vice versa requires deeper investigation to embed these influences in the models.

## 3  APPROACH

### 3.1  Mutual Influence of Trust: Trust Cycles

The goal of our research is to develop a method that can be used by the robot in order to continuously estimate the human trust in it and form its own trust toward the human and adapt its behavior accordingly. For this, we use a model proposed in [6], which represents a one-way trust cycle between a human and intelligent agent (the cycle with solid blue arrows in figure 1) and does not include the mutuality of the concept. This cycle shows how an autonomous intelligent system (a robot in our case) provides assurances to the human and tries to detect the trust-related behavior of the human. We modify this model by adding the reversed direction to it. In this direction, the human also detects the trust-related behavior of the robot and changes his/her behavior to provide assurances that the intuitive common understanding still the basis of joint actions. Figure 1 shows a diagram of the modified model that represents the resulting trust cycles between humans and robots.

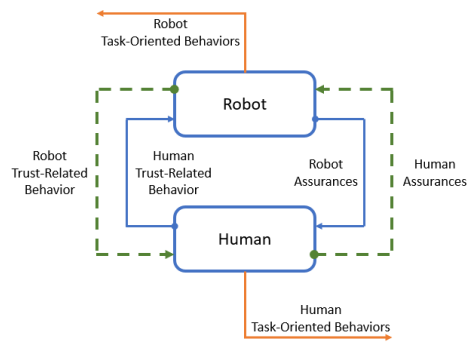

Figure 1: Mutual trust cycles between a human and a robotic partner in a physical collaboration setting. Solid lines show the original model by [6], dashed lines show the additions made by the authors of this paper.

Complementing the original model, trust-related behavior of the robot may include forms of conservative behaviors (e.g. force or speed reduction) caused by low trust, and switching back to normal when trust is in an appropriate level. It should be noted here, that unlike human interpersonal trust where a higher partner trust may affect the other partner's trust positively [18], in HRC the opposite could be true quite often: In many cases, a high trust level can be associated with overreliance of the human on the robot, which can be associated with risky or irrational behavior. In such cases, high human trust should decrease the trust of the robot toward the human. Building a model of mutual trust formation between humans and robots therefore needs careful analysis of tasks and human behavior. The assurances from human side should be intuitive and task-related, such as: the fulfillment of the pre-defined conditions of a task, human attention, and adherence to safety requirements. We hypothesize that such assurances do not pose additional workload on the human since they already exist.

Our approach for designing and implementing this model consists of four necessary steps, they are: (1) Identifying the relevant trust dimensions in physical hand-in-hand collaboration settings, where the partners work in close proximity and each agents' actions are interdependent. (2) Developing a model of human trust in the robot using the identified trust dimensions. (3) Developing a method for building the robot trust in the human partner without compromising human autonomy and increasing the workload. (4) Combining both models and creating a computational model that includes them both.

### 3.2 Implementation Steps

In order to identify the aforementioned dimensions of trust, we aim at conducting an experiment in which human participants will collaborate with a collaborative robot on the disassembly of a model of an electric car battery, which consists of many modules fixed with screws. This first experiment is going to take place in a mixed-reality environment to eliminate the risk of non-expert participants collaborating with a real robot. The environment setup for this experiment is now being designed (Appendix A) and a detailed plan on the study is being prepared.

During the experiment, the human task is to position the battery somewhere accessible by the robot, assuming that the human has the ability to lift the battery, and to make sure that the battery stays in this position and orientation during the robot operation. The robot task is to help the human by loosening the screws for him/her. The propensity of each participant to trust a robotic partner in this setting will be identified to avoid any biases in the model and to personalize it. After the experiment, a survey is to be filled that contains all factors that exist in the literature to find the relevant dimensions in this physical collaboration setting.

The next step after the experiment is to derive a computational model that incorporates the identified trust dimensions, which helps the robot adapt its behavior based upon. The validity of the developed model will be checked in a second similar experiment with real robot. Afterwards, the computational trust model from the robot side can be formed which should take the influence of the human trust level into consideration. For this model, the robot needs to monitor the environment including human behavior and always check for anomalies that might compromise the human safety as primary goal and task performance as secondary one. We also emphasize the difference between the human trust model and the robot trust model by considering the objectives of the robot as the main components of the model.

### 4 CONCLUSION

In this paper we proposed a framework for mutual trust modeling for successful human-robot collaboration. Our present trust modeling process calculate for the differences between humans and robots. The human trust model incorporates all trust factors that affect human trust in a robot, whereas the robot trust model focuses more on human safety as primary motivation for it and task performance as a secondary one. Combining and implementing these models in the robot have the potential to make the interaction more natural and human-like. We also believe that this does not reduce the human autonomy but rather enhances it since the robot always considers the human trust and safety as bases of decision making and action selection processes.

**APPENDIX A: EXPERIMENT SETUP**

The collaborative robot, Panda, is simulated on a Microsoft HoloLens 2 head mounted device will support humans by loosening screws for them in a Mixed-Reality environment.

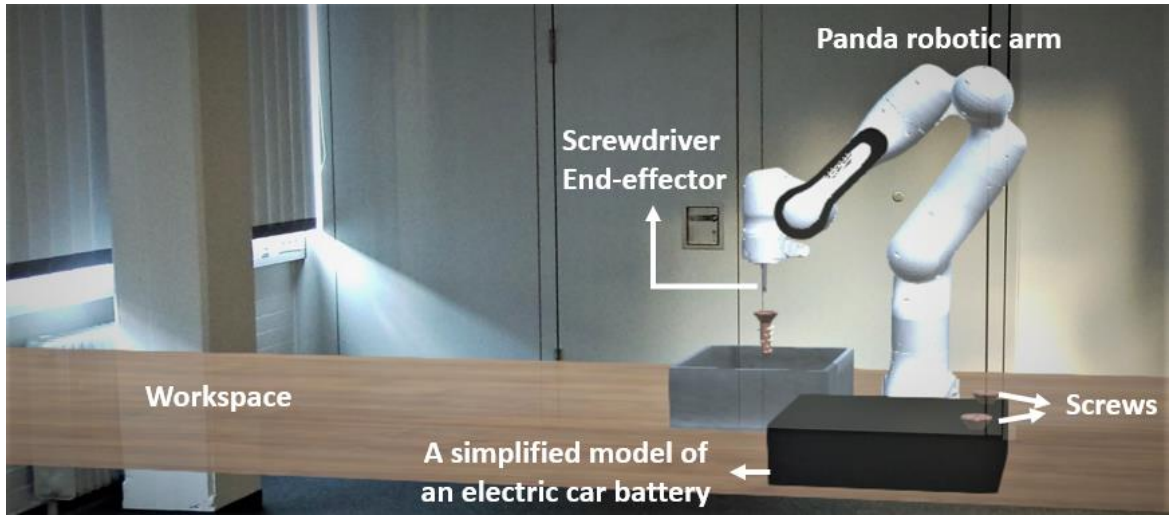

Figure 2: Experiment setup in Mixed-Reality environment